\documentclass[letterpaper, 10 pt, conference]{ieeeconf}  

\IEEEoverridecommandlockouts                              

\usepackage{amsmath} 
\usepackage{amssymb}  
\usepackage{graphicx}
\usepackage{booktabs}
\usepackage{subcaption}
\usepackage{url}
\usepackage{balance}
\title{\LARGE \bf
Floor Friction Assessment from Robot Tire-Surface Interaction
}

\author{Kulbhushan Komarraju$^{1}$, Nathaniel Pearson$^{2}$,  Yu Gu$^{1}$, and Jason Gross$^{1}$
\thanks{$^{1}$ Department of Mechanical, Materials and Aerospace Engineering, West Virginia University}
\thanks{$^{2}$Lane Department of Computer Science and Electrical Engineering, West Virginia University}%
\thanks{This work was supported in part by NSF Projects \#2132937 and \#2132936  and the NASA West Virginia Space Grant Consortium, grants \#80NSSC20M0055 and \#80NSSC25M7079.}
}

\begin{document}

\maketitle
\thispagestyle{empty}
\pagestyle{empty}

\begin{abstract}
Floor traction influences wheel-slip behavior, dead-reckoning accuracy, and the overall navigation performance of wheeled mobile robots that rely on wheel odometry. As these robots become increasingly common in GNSS-denied indoor environments, including automated manufacturing and retail facilities, their onboard measurements could also be used to characterize floor conditions and improve localization. In this study, a skid-steer robot was driven over five floor surfaces with distinct traction characteristics using commanded velocity profiles designed to induce wheel slip through repeated acceleration and direction reversal. Differences between the body velocity estimated from inertial sensing and the velocity obtained from wheel odometry were evaluated using the integrated relative-velocity error, phase-derived wheel-to-body response delay, and velocity-reversal response delay. Welch power spectral density analysis was additionally applied to characterize the frequency distribution of the velocity mismatch during positive and negative acceleration. Band-integrated spectral power, spectral centroid, and spectral decay rate were extracted from each recording and evaluated using linear discriminant analysis and multiclass classification. The response-delay measurements produced consistent surface-dependent trends, with the longest delays observed on soap and the shortest delays observed on carpet. The spectral features also provided clear separation among the tested surfaces, achieving a balanced classification accuracy of 97.1\% under recording-level cross-validation. These results demonstrate that both the magnitude and frequency domain characteristics of wheel-to-body velocity mismatch contain useful information about tire-surface interactions. Such information could support surface recognition and the development of slip-aware wheel-odometry weighting within robot state-estimation frameworks.
\end{abstract}

\section{INTRODUCTION}

When wheeled mobile robots (WMRs) traverse on a surface, they experience slip depending on the traction between the floor and the tires \cite{c1}. Tire-surface slip can be detrimental to wheel-odometry-based dead-reckoning accuracy for WMRs operating in GNSS-denied indoor environments \cite{c2} \cite{c3}. Wheel odometry can become biased due to slip, leading to drift in dead-reckoning and degraded state estimates. This problem is especially relevant for skid-steer robots, where turning, by design, induces lateral and longitudinal slip (i.e., skid) that is strongly dependent on surface traction.

A practical way to characterize floor traction is through the coefficient of friction (CoF), but CoF is rarely known \textit{a priori} in real deployments and can vary across space and time due to changes in flooring material, cleaning practices, or contamination. Rather than treating slip as an unmodeled disturbance, a navigation-centric approach is to infer traction-related information online and incorporate it into state estimation, enabling slip-aware dead reckoning and more reliable motion prediction \cite{c4} \cite{c5}.

In this work, we propose a slip-based floor slipperiness estimation method using a skid-steer robot equipped with an IMU and wheel encoders, together with an exteroceptive odometry source. A preliminary assessment of the approach is conducted in which the robot is driven on multiple surfaces with different friction properties using a velocity command profile designed to intentionally induce skidding. A slip velocity signal is computed as the mismatch between the robot’s body velocity and wheel-odometry-derived velocity. In addition to measuring the magnitude of this velocity mismatch, we evaluate the wheel-to-body velocity response using a phase-derived response delay and a velocity-reversal response delay. Frequency-domain features derived from robot-terrain interactions have previously been used for terrain characterization and classification\cite{c6} \cite{c7}. Hence, for further analysis, the relative-velocity signal is also analyzed in the frequency domain, and spectral features are extracted to determine whether the surfaces produce distinguishable frequency-domain signatures. To obtain a body-velocity estimate, we fuse IMU measurements with exteroceptive velocity using a factor-graph estimator. For this preliminary work, as a starting point, we use motion-capture estimates as a proxy for exteroceptive odometry, such as visual or LiDAR odometry. The resulting slip, response-delay, and spectral signatures are shown to be surface-dependent and provide complementary information about floor traction.

Our results demonstrate clear separation in slip behavior across five indoor surfaces, and we show a strong correlation between a simple slip metric and ground-truth friction measurements. The phase-derived and velocity-reversal response delays also show consistent surface-dependent trends. In addition, the extracted spectral features distinguish the five surfaces and achieve a balanced classification accuracy of 97.1\% under recording-level cross-validation. These findings suggest that tire-surface interaction signatures can be leveraged to infer floor slipperiness and identify the surface being traversed.

The remainder of this paper is organized as follows. Section II reviews related work on friction estimation, slip modeling, and slip-aware state estimation for mobile robots. Section III describes the formulation used to compute body velocity and defines the slip-velocity, wheel-to-body response-delay, and spectral metrics obtained by comparison with wheel-odometry velocity. Section IV presents the experimental platform, sensor stack, and data collection procedure. Section V reports experimental results and compares the slip, body velocity response delay, and spectral signatures across surfaces. Section VI concludes the paper and outlines directions for future work.

\section{BACKGROUND}



\subsection{Friction Estimation using Robots}

Robotic friction estimation has been studied across several domains, with methods typically tied to the robot’s interaction modality and sensing stack. In robotic manipulation, for example, friction is often estimated using tactile sensing and incipient-slip detection, where local contact events are used to infer friction limits and adjust grasp forces \cite{c8}\cite{c9}. Classic and subsequent work in this area demonstrates that friction-related information can be recovered from slip-onset signatures at the contact interface.

In legged locomotion, friction estimation is important for stable foothold selection and slip-aware control, especially on low-traction or changing terrain. These methods commonly estimate contact state, slip, or terrain parameters online to improve locomotion robustness \cite{c10}\cite{c11}. They typically rely on proprioceptive sensing (e.g., IMU, foot pressure sensors and joint measurements) and infer slip through contact-consistent state estimation, where violations of no-slip contact assumptions indicate traction loss.

For WMRs, friction and traction are frequently studied through wheel slip ratio and wheel-ground interaction models \cite{c12}. Prior work has addressed slip estimation for localization and control, and some studies explicitly estimate friction-related parameters from vehicle motion behavior \cite{c13}. These approaches establish that wheel-body velocity mismatch contains useful information about surface interaction \cite{c14}\cite{c15}, but many are developed for outdoor mobility, deformable terrain traversal, or control objectives rather than indoor navigation settings where floor traction varies across common building materials and has an additional challenge of lack of GNSS for accurate localization.

\subsection{Sensor Fusion for Slip Estimation}

Reliable slip estimation requires an accurate estimate of the robot’s state in order to compare body and wheel velocities. IMUs are widely available and provide high-frequency motion measurements, but inertial navigation system (INS) integration errors accumulate over time, leading to drift without external corrections. As a result, fusing INS with exteroceptive measurements using probabilistic estimators such as Kalman filters, particle filters, or factor-graphs is common practice \cite{c16}. Sensor fusion improves accuracy and robustness by leveraging complementary strengths of different sensors and reducing the impact of noise, drift, and bias from any single source \cite{c17}. The resulting state estimates can then be compared against wheel odometry to infer slip characteristics that are informative of traction \cite{c18}.

\subsection{Slip-Based Friction Inference for Skid-Steer Robots}

Skid-steer robots are especially well-suited for friction-sensitive analysis because turning and aggressive maneuvers directly induce lateral and longitudinal slip. Unlike ideal rolling kinematics, skid-steer motion intrinsically depends on tire-floor interaction \cite{c19}. This causes the measured body velocity to diverge from wheel-inferred kinematics in a manner that depends on surface traction. This creates an opportunity to use slip velocity (i.e., the mismatch between body and wheel velocities) as a compact descriptor of floor slipperiness.

This paper differs from prior robotic friction-estimation studies in two important ways. First, rather than using tactile sensors or direct force measurements, it infers friction-related behavior from vehicle-level slip patterns obtained from standard mobile-robot sensing. Second, it targets traction-aware indoor navigation by using repeatable slip signatures under controlled maneuvers to infer floor friction. The floor traction is assessed through complementary analyses of the wheel and body velocity recordings collected during controlled back-and-forth maneuvers. Body velocity is estimated independently of wheel odometry by fusing inertial and motion-capture measurements in a factor graph. The difference between these velocities is then used to characterize wheel slip. Spectral features extracted from Welch power spectra describe how the slip signal is distributed across frequencies and are used to distinguish the tested surfaces. The magnitude of the velocity mismatch is examined using the cumulative slip, while its timing is examined through phase-derived delays and differences in velocity zero-crossing times. These magnitude and delay measures are compared with independently measured coefficients of friction to examine their relationship with floor traction.  Future work will consider integrating this traction information into state estimation to improve dead-reckoning performance in indoor navigation tasks in degraded environments (e.g., dark, dusty, or foggy) that would cause problems in visual or LiDAR-based approaches.

\section{METHODOLOGY}

\subsection{Robotic System Overview}
For this paper, a skid-steer mobile robot was used to estimate surface-dependent slip characteristics through analysis of the wheel slip, that is, the relative velocity between wheel-derived odometry and the robot's actual body velocity. A practical challenge is that body velocity estimates can be noisy or biased when derived from a single sensing source. In this work, we address this by using a factor graph to fuse motion capture position and velocity estimates (e.g., Vicon) and IMU data, yielding a temporally consistent estimate of robot body velocity. Motion-capture systems are not commonly available in real-world deployments, so in this preliminary study we use MoCap only to validate the proposed friction estimation methodology. As a next step, MoCap would be replaced with more deployable exteroceptive sources such as visual odometry or LiDAR odometry, fused with inertial measurements for body-velocity estimation. The fused body velocity is then compared with wheel odometry-derived velocity to quantify relative motion and extract slip patterns across surfaces. This formulation emphasizes state-estimation quality as a key enabler of friction inference.

\begin{figure}[t]
    \centering
    \includegraphics[width=\columnwidth]{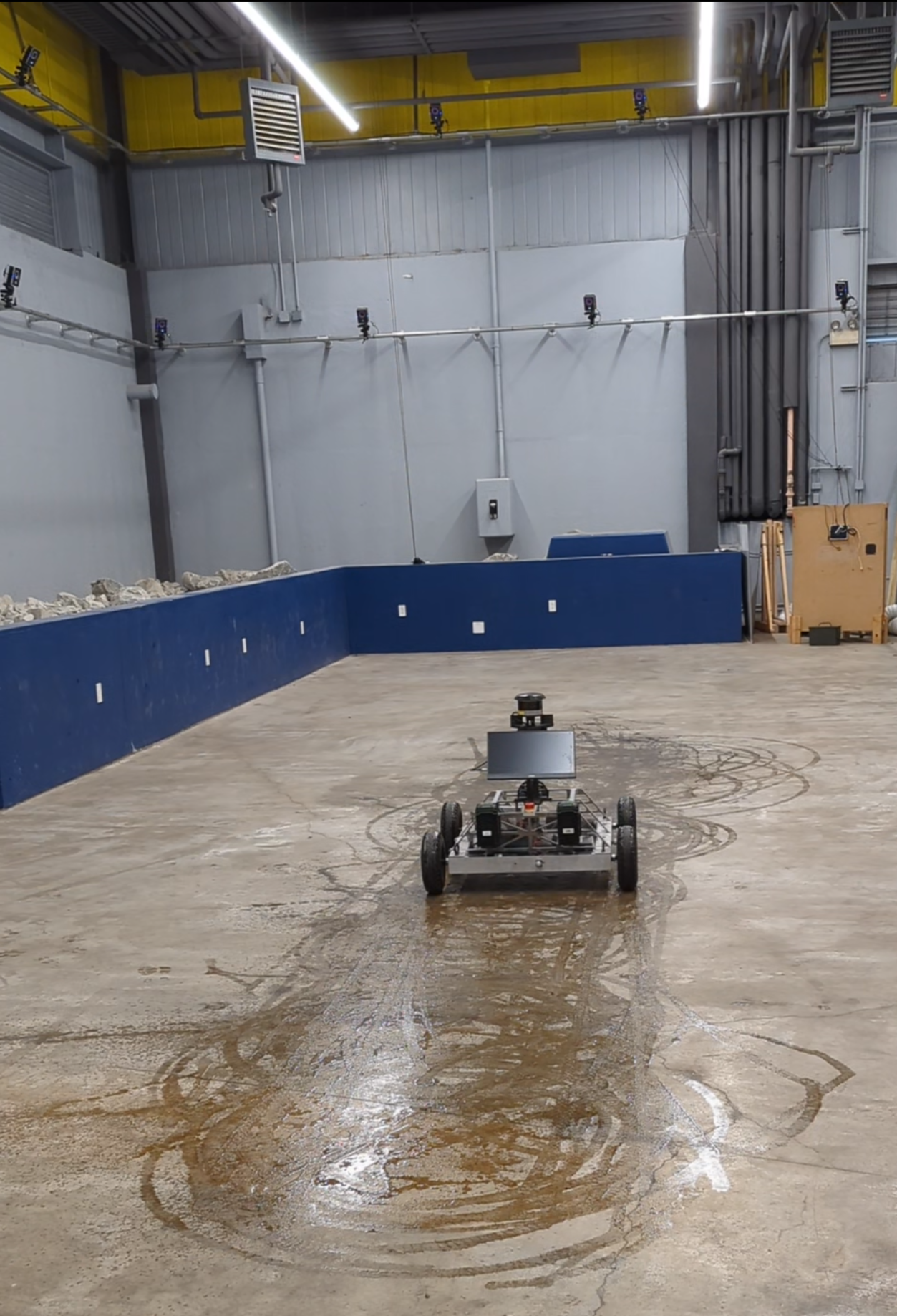}
    \caption{Skid-steer robot undergoing experiments in the motion capture facility. In this example experiment, the driving surface is covered with a soapy mixture. }
    \label{fig:Vicon}
\end{figure}

The objective is to estimate body velocity independently of wheel measurements and compute slip-induced relative velocity to assess floor slipperiness.

\subsection{Factor-Graph State Estimation}

\begin{figure}
    \centering
    \includegraphics[width=0.8\linewidth]{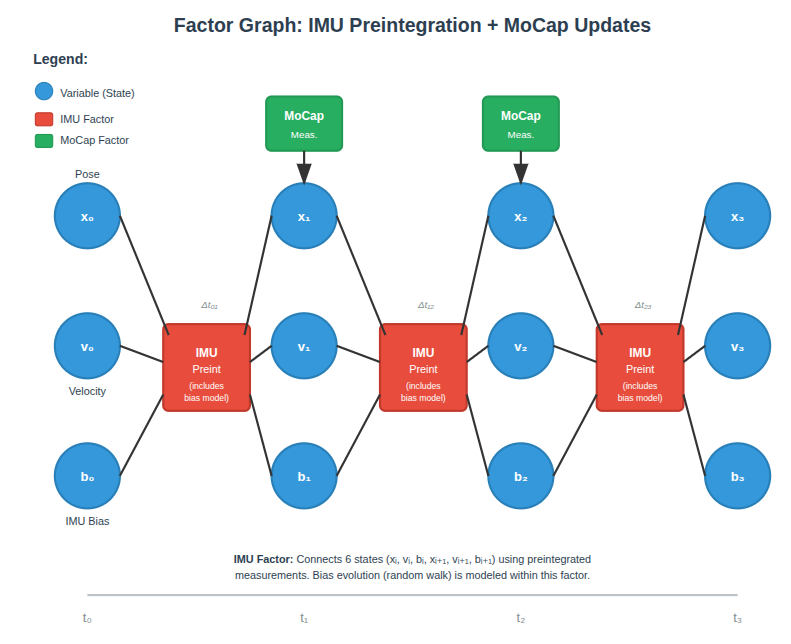}
    \caption{Schematic of the factor graph estimation being used for robot state estimation. Blue nodes represent the state variables (pose, velocity, and IMU biases) at each timestep. Red factors denote IMU preintegration constraints between consecutive states, while green factors represent Motion Capture (MoCap) pose and velocity measurement updates.}
    \label{fig:FG}
\end{figure}

Sensor data from the IMU and Motion capture (Mocap) system are fused using an Incremental Smoothing and Mapping (ISAM2) factor graph framework~\cite{c20}. IMU measurements accumulate drift with time and Mocap velocity measurements are noisy. The sensor fusion approach provides improved pose and velocity estimates compared to individual sensor measurements. A schematic of the factor graph is shown in Fig.\ref{fig:FG}.

Robot states at timestep $k$ are represented as nodes:

\begin{equation}
\mathbf{X}_k = \{\mathbf{T}_k, \mathbf{v}_k, \mathbf{b}_k\}
\end{equation}

where:
\begin{itemize}
    \item $\mathbf{T}_k \in SE(3)$ is the robot pose, represented as a 7$\times$1 vector
\[
\mathbf{T}_k =
\begin{bmatrix}
x_k & y_k & z_k & q_{w,k} & q_{x,k} & q_{y,k} & q_{z,k}
\end{bmatrix}^\top,
\]
where $(x_k,y_k,z_k)$ is the 3-D position and $(q_{w,k},q_{x,k},q_{y,k},q_{z,k})$ is the unit quaternion orientation.
\item $\mathbf{v}_k \in \mathbb{R}^3$ is the body (linear) velocity expressed in the world frame,
\[
\mathbf{v}_k =
\begin{bmatrix}
v_{x,k}\\ v_{y,k}\\ v_{z,k}
\end{bmatrix}.
\]

\item $\mathbf{b}_k$ represents IMU biases, split into accelerometer and gyroscope biases:
\[
\mathbf{b}_k =
\begin{bmatrix}
\mathbf{b}^a_k\\ \mathbf{b}^\omega_k
\end{bmatrix},
\quad
\mathbf{b}^a_k =
\begin{bmatrix}
b^a_{x,k}\\ b^a_{y,k}\\ b^a_{z,k}
\end{bmatrix},
\quad
\mathbf{b}^\omega_k =
\begin{bmatrix}
b^\omega_{x,k}\\ b^\omega_{y,k}\\ b^\omega_{z,k}
\end{bmatrix}.
\]
\end{itemize}

Sensor measurements are modeled as probabilistic constraints (factors) between nodes.

\subsubsection{IMU Preintegration}

High-frequency IMU measurements are pre-integrated using the standard pre-integrated IMU model:

\begin{equation}
\mathbf{X}_{k} = f(\mathbf{X}_{k-1}, \mathbf{u}_{\text{IMU}}, \mathbf{b}_{k-1})
\end{equation}

where $\mathbf{u}_{\text{IMU}}$ denotes tri-axial orthogonal accelerometer and gyroscope measurements.

Preintegration accumulates motion constraints at 125 Hz, enabling incorporation of inertial acceleration and angular velocity of the robot body for kinematic estimates \cite{c21}.

\subsubsection{Mocap Pose and Velocity Update}

Motion capture pose measurements at 192 Hz provide world-frame pose and velocity priors that correct drift accumulated in uncorrected body-frame inertial navigation. After each preintegration step, the Mocap update with the closest timestamp is processed in the factor graph. The pose updates are given much higher weight than velocity updates as Mocap pose estimates are relatively accurate while the velocity estimates are often noisy as the velocity is derived from its pose estimates by time differentiation. These are modeled as measurement constraints:

\begin{align}
\mathbf{T}_k &\sim \mathcal{N}(\mathbf{T}_{\text{Mocap}}, \Sigma_T)\\
\mathbf{v}_k &\sim \mathcal{N}(\mathbf{v}_{\text{Mocap}}, \Sigma_v)
\end{align}


\subsection{Robot Wheel Slip Estimation}

Because odometry and factor graph states are sampled at different frequencies, timestamp alignment is performed by associating each odometry timestamp with the nearest factor graph estimate.

Let:
\begin{itemize}
    \item $v_{\text{odom}}$ denotes the longitudinal wheel velocity,
    \item $\mathbf{v}_{\text{FG}}$ denotes the factor graph velocity in world-frame.
\end{itemize}

The longitudinal body velocity $v_{\text{body}}$ is computed using the estimated yaw angle $\psi$:
\begin{equation}
v_{\text{body}} = v_x \cos(\psi) + v_y \sin(\psi)
\end{equation}

where $v_x$ and $v_y$ are the x and y components of $\mathbf{v}_{\text{FG}}$. The relative, or slip, velocity is then defined as:
\begin{equation}
v_{\text{rel}} = v_{\text{odom}} - v_{\text{body}}
\end{equation}

This quantity estimates the mismatch between ideal pure-rolling wheel kinematics and true body motion.

\subsection{Signal Conditioning and Slip Quantification}

The relative velocity segments are categorized into up and down segments based on the slip velocity profile going from the negative maximum to the positive maximum (acceleration phase) or vice versa. This is done to improve clarity and to use each of these segments as an individual data point to calculate the median Area Under the Curve (AUC). For each segment, the unsigned AUC is computed:


\begin{equation}
\text{AUC}_{\text{abs}} = \int_{t_s}^{t_e} |v_{\text{rel}}(t)| \, dt
\end{equation}

This metric represents the cumulative slip magnitude independent of direction. $t_s$ and $t_e$ denote the start and end times of each segment. Segment-wise AUC values are averaged per surface category for comparison.



\subsection{Relative-Velocity Spectral Distribution}

The frequency distribution of the wheel-to-body velocity mismatch was also
examined using the relative velocity. Positive- and
negative-slope segments of the body-velocity signal were analyzed and each segment was resampled onto a uniform time grid using
its median sampling interval. The power spectral density,
$S_{e,i}(f)$, of segment $i$ was estimated with a 50\% window overlap. 

The relative-velocity power spectral density was estimated using
Welch's method, in which windowed periodograms from overlapping signal
segments are averaged to reduce the variance of the spectral estimate
\cite{c22}. A Hann window was applied to reduce spectral leakage caused
by the finite duration of each velocity segment \cite{c23}. The phase
difference between the wheel-odometry and body-velocity signals was
converted into an equivalent response delay following established
frequency-domain delay-estimation principles \cite{c24}.

To compare spectral shape independently of the total mismatch power,
each spectrum was normalized according to

\begin{equation}
S_{e,i}^{\mathrm{norm}}(f)
=
\frac{S_{e,i}(f)}
{\displaystyle
\int_{f_{\min}}^{f_{\max}}
S_{e,i}(\nu)\,d\nu},
\label{eq:normalized_psd}
\end{equation}

where $f_{\min}=0.25$ Hz and $f_{\max}=15$ Hz. The normalized spectra
were converted to decibels using

\begin{equation}
S_{e,i}^{\mathrm{norm,dB}}(f)
=
10\log_{10}
\left[
S_{e,i}^{\mathrm{norm}}(f)
\right].
\label{eq:normalized_psd_db}
\end{equation}

Frequencies below the intrinsic resolution limit, $1/T_i$, were
excluded for each segment, where $T_i$ is the segment duration.
Frequencies above the corresponding Nyquist frequency were also
excluded. The remaining spectra were interpolated onto a common
frequency grid, and the median was calculated at frequencies containing
at least three valid segments.

To determine whether the relative-velocity spectra contained sufficient
information to distinguish the surfaces, three features were extracted
from the run-level mean PSD over the common frequency interval from
$3.5$ to $9.5$ Hz based on the segment-dependent frequency resolution and the Nyquist frequency respectively. The features were calculated separately for the negative and positive FG slopes. Recordings containing fewer than four valid slope segments in either direction were excluded because an insufficient number of segments could produce an unstable mean PSD dominated by individual transitions or measurement noise.

For recording $r$ and slope direction $d$, the integrated spectral
power was calculated as

\begin{equation}
E_{r,d}
=
10\log_{10}
\left[
\int_{f_{\mathrm{L}}}^{f_{\mathrm{H}}}
P_{r,d}(f)\,df
\right],
\label{eq:spectral_energy}
\end{equation}

where $P_{r,d}(f)$ is the run-level mean PSD,
$f_{\mathrm{L}}=3.5$ Hz, and $f_{\mathrm{H}}=9.5$ Hz. The spectral
centroid was calculated as

\begin{equation}
f_{\mathrm{c},r,d}
=
\frac{
\displaystyle
\int_{f_{\mathrm{L}}}^{f_{\mathrm{H}}}
fP_{r,d}(f)\,df
}{
\displaystyle
\int_{f_{\mathrm{L}}}^{f_{\mathrm{H}}}
P_{r,d}(f)\,df
}.
\label{eq:spectral_centroid}
\end{equation}

The rate at which the PSD decreased with frequency was determined by
fitting

\begin{equation}
10\log_{10}\left[P_{r,d}(f)\right]
=
m_{r,d}f+b_{r,d},
\label{eq:spectral_slope}
\end{equation}

where $m_{r,d}$ is the PSD slope in dB/Hz. Each recording was therefore
represented by the six-feature vector

\begin{equation}
\mathbf{x}_r
=
\left[
E_{r,-},
E_{r,+},
f_{\mathrm{c},r,-},
f_{\mathrm{c},r,+},
m_{r,-},
m_{r,+}
\right]^{\mathrm{T}}.
\label{eq:spectral_feature_vector}
\end{equation}

A shrinkage linear discriminant analysis was used to
project the six-dimensional feature vectors onto two discriminant axes
for visualization. Because LDA uses the known class labels when
determining these axes, this projection was treated as descriptive and
was not used as the validation result.

Surfaces were classified using a multiclass logistic regression model with class balancing. The model was trained by leaving out one recording for cross validation and tested. Features were standardized using the mean and standard deviation of the training recordings in each cross-validation fold. The same scaling was then applied to the recording left out for testing. Classification performance was reported using balanced accuracy.

\begin{equation}
\mathrm{BA}
=
\frac{1}{C}
\sum_{c=1}^{C}
\frac{\mathrm{TP}_{c}}
{\mathrm{TP}_{c}+\mathrm{FN}_{c}},
\label{eq:balanced_accuracy}
\end{equation}

where $C$ is the number of surfaces.

\subsection{Frequency-Domain Velocity Response Analysis}

The AUC-based analysis quantifies the accumulated magnitude of the 
difference between the wheel-odometry velocity and the ground-referenced 
body velocity. However, it does not distinguish an amplitude mismatch 
from a temporal delay between the two signals. A complementary 
frequency-domain analysis was therefore performed to quantify how 
rapidly the body velocity followed the odometry velocity during repeated 
direction reversals.
Each complete recording was treated as one independent observation. The
primary excitation frequency was determined from the median time between
consecutive command reversals. Because two reversals represent one
complete oscillation, the reversal frequency was calculated as

\begin{equation}
f_0 =
\left[
2\,\mathrm{median}_{k}
\left(
t_{\mathrm{r},k+1}-t_{\mathrm{r},k}
\right)
\right]^{-1},
\label{eq:reversal_frequency}
\end{equation}

where $t_{\mathrm{r},k}$ is the time of the $k$th command reversal.

The Fourier component of each velocity signal was evaluated at the exact
reversal frequency. This avoided restricting the analysis to the nearest
discrete FFT frequency bin. The odometry and body-velocity signals were
represented using the sinusoidal least-squares model

\begin{equation}
\begin{aligned}
v_x(t) ={}& a_x + b_x t
+ c_x \cos(2\pi f_0 t) \\
&+ d_x \sin(2\pi f_0 t),
\qquad x \in \{\mathrm{o},\mathrm{b}\},
\end{aligned}
\label{eq:harmonic_fit}
\end{equation}

where $a_x$ and $b_x$ account for the signal offset and slow linear drift,
respectively. The complex Fourier coefficient at $f_0$ was then defined
as

\begin{equation}
V_x(f_0) = c_x - j d_x.
\label{eq:complex_coefficient}
\end{equation}

The frequency-response function relating the body velocity to the
wheel-odometry velocity was calculated as

\begin{equation}
H_{\mathrm{bo}}(f_0)
=
\frac{V_{\mathrm{b}}(f_0)}
     {V_{\mathrm{o}}(f_0)}
=
G(f_0)\exp\left[j\phi(f_0)\right],
\label{eq:transfer_function}
\end{equation}

where $G(f_0)$ is the body-to-odometry gain and $\phi(f_0)$ is the phase
difference between the signals. The phase difference was converted to
an equivalent frequency-domain catch-up delay according to\cite{c24}

\begin{equation}
\tau_{\phi}
=
-\frac{\phi(f_0)}
       {2\pi f_0}.
\label{eq:phase_delay}
\end{equation}

A positive value of $\tau_{\phi}$ indicates that the body velocity lags
the wheel-odometry velocity. Thus, $\tau_{\phi}$ represents the effective
delay with which the translational motion of the robot follows the wheel
response at the primary reversal frequency.

A direct time-domain measurement was also calculated to validate the
frequency-domain delay. For each direction reversal, the zero-crossing
times of the odometry and body-velocity signals were determined using
linear interpolation between the samples immediately before and after
the crossing. The direct reversal velocity response delay was calculated as

\begin{equation}
\tau_0
=
t_{\mathrm{b},0}
-
t_{\mathrm{o},0},
\label{eq:zero_crossing_delay}
\end{equation}

where $t_{\mathrm{o},0}$ and $t_{\mathrm{b},0}$ are the zero-crossing
times of the odometry and body-velocity signals, respectively. A
positive value of $\tau_0$ indicates that the body changed direction
after the wheel-odometry signal.

Multiple direction reversals were present within each recording.
Therefore, the median zero-crossing delay across all valid reversals was
calculated for each recording before comparing surfaces. This prevented
individual reversals from being treated as independent experimental
replicates. The frequency-domain catch-up delay, $\tau_{\phi}$, and the
direct reversal catch-up delay, $\tau_0$, were subsequently used as the
two measures of wheel-to-body response delay.

\section{Experimental Setup}

\subsection{Hardware Platform}


Experiments were conducted using a skid-steer mobile robot equipped with an IMU and encoders for proprioceptive sensing. The robot's weight is measured as 40.3 kg and its tires are made of rubber (what exact material?). An infrared motion capture system has been used as an independent source of robot's pose and velocity measurements. The sensors used are summarized below:
\begin{itemize}
    \item An ADIS-16445 Analog Devices Inertial Measurement Unit (IMU) operating at 125~Hz,
    \item A Vicon Mocap providing pose and velocity measurements at 192~Hz and,
    \item Wheel encoders with a resolution of $0.222^\circ$ and rate of 20~Hz.
\end{itemize}

\subsection{Surface Conditions}

Experiments were performed on five surfaces with different friction characteristics:

\begin{itemize}
    \item Carpet
    \item Concrete
    \item Melamine board
    \item Soap-oil surface (referred to as Soap surface hereafter)
    \item Vinyl sheet
\end{itemize}

These surfaces span varying friction conditions, with the soap surface being the most slippery and the carpet the least slippery.

\subsection{Testing Procedure}

\begin{figure}
    \centering
    \includegraphics[width=0.8\linewidth]{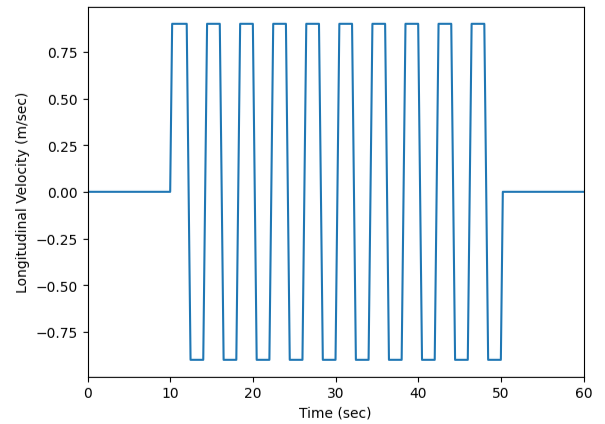}
    \caption{Commanded velocity profile}
    \label{fig:cmd_vel}
\end{figure}

For each surface, the robot executed repeated forward–reverse maneuvers at its maximum longitudinal speed, with direction changes occurring every 2 s (4 s per full cycle). At each reversal, the commanded velocity was switched instantaneously to induce rapid braking and direction reversal; the corresponding command profile is shown in Fig.~\ref{fig:cmd_vel}. The resulting rapid acceleration and deceleration at the start and end of each segment intentionally induced longitudinal slip. For each test configuration, five independent trials were recorded.

All sensor data were logged with preserved timestamps for offline processing.

\subsection{Evaluation Metric}

One of the metrics reported in this study is the mean segment-wise unsigned AUC of relative velocity for each surface category. For some of the trials, the duration of the command velocity profile is not identical as others. To remove any bias, all AUC values are normalized by their respective time durations. Larger AUC values indicate increased cumulative slip and correspond to lower friction conditions.

\section{RESULTS}

\begin{figure}[t]
    \centering
\begin{subfigure}{0.48\columnwidth}
    \centering
    \includegraphics[width=\linewidth]{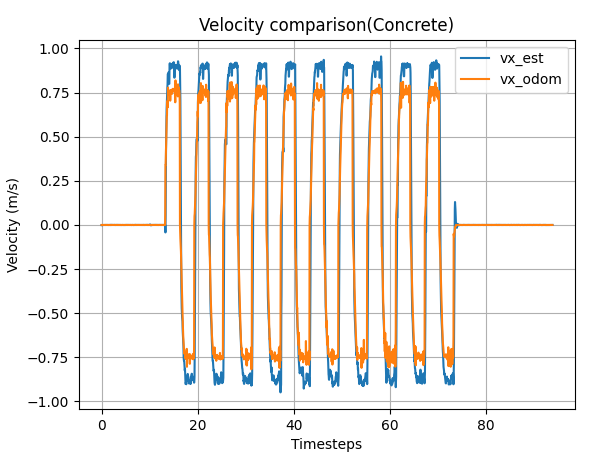}
\end{subfigure}
\hfill
\begin{subfigure}{0.48\columnwidth}
    \centering
    \includegraphics[width=\linewidth]{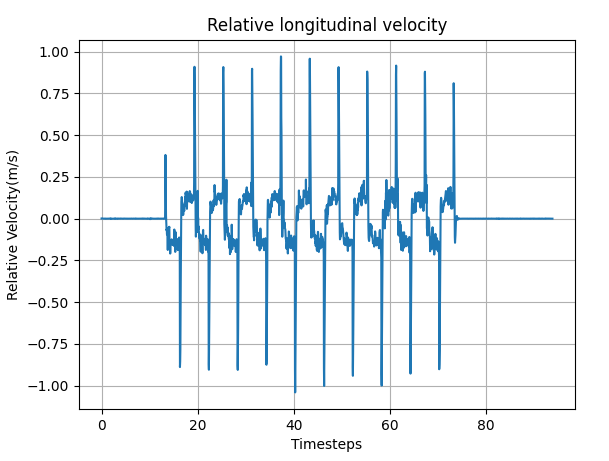}
\end{subfigure}
\caption{Velocity comparison and relative velocity profile for Concrete surface.}
\label{fig:Concrete}
\end{figure}

\begin{figure}[t]
    \centering
\begin{subfigure}{0.48\columnwidth}
    \centering
    \includegraphics[width=\linewidth]{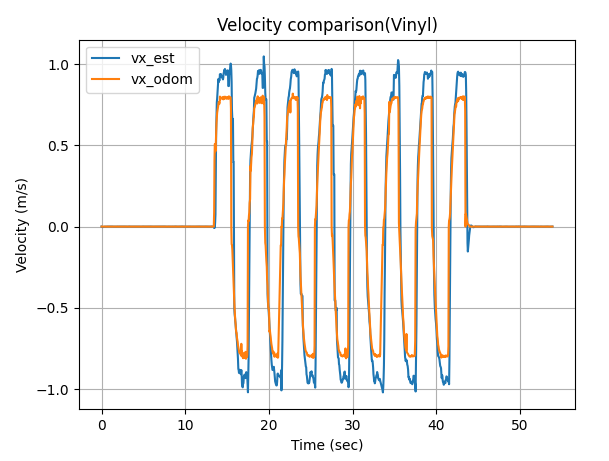}
\end{subfigure}
\hfill
\begin{subfigure}{0.48\columnwidth}
    \centering
    \includegraphics[width=\linewidth]{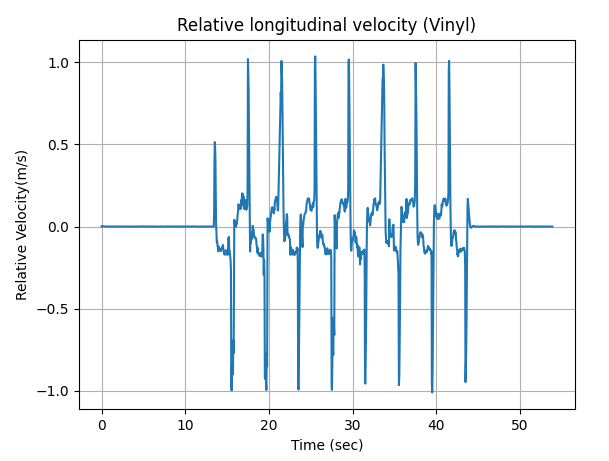}
\end{subfigure}
    \caption{Velocity comparison and relative velocity profile for Vinyl surface.}
    \label{fig:Vinyl}
\end{figure}

\begin{figure}[htb!]
\begin{subfigure}{0.48\columnwidth}
    \centering
    \includegraphics[width=\linewidth]{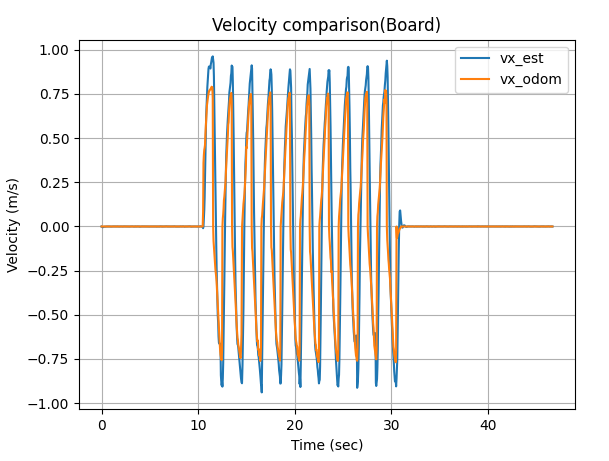}
\end{subfigure}
\hfill
\begin{subfigure}{0.48\columnwidth}
    \centering
    \includegraphics[width=\linewidth]{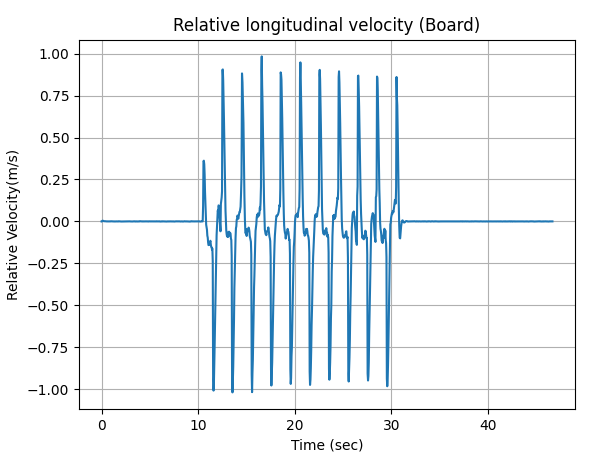}
\end{subfigure}
    \caption{Velocity comparison and relative velocity profile for Melamine Board surface.}
    \label{fig:Board}
\end{figure}

\begin{figure}[htb!]
\begin{subfigure}{0.48\columnwidth}
    \centering
    \includegraphics[width=\linewidth]{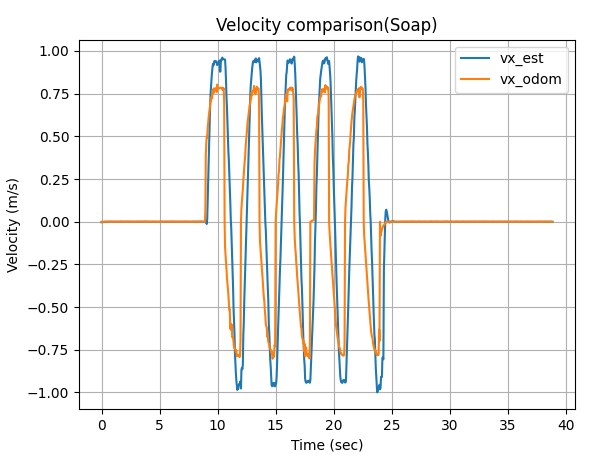}
\end{subfigure}
\hfill
\begin{subfigure}{0.48\columnwidth}
    \centering
    \includegraphics[width=\linewidth]{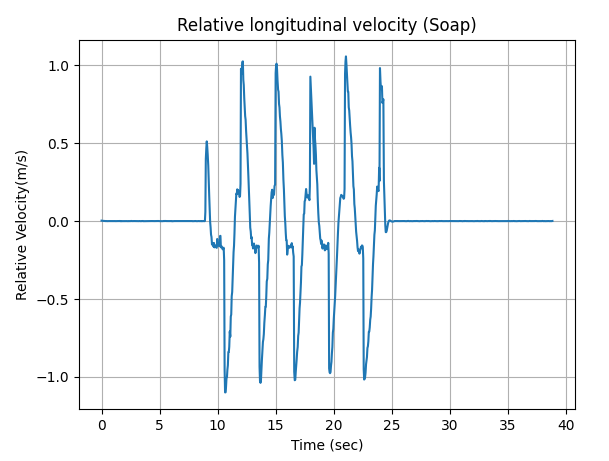}
\end{subfigure}
    \caption{Velocity comparison and relative velocity profile for Soap surface.}
    \label{fig:Soap}
\end{figure}

\begin{figure}[htb!]
\begin{subfigure}{0.48\columnwidth}
    \centering
    \includegraphics[width=\linewidth]{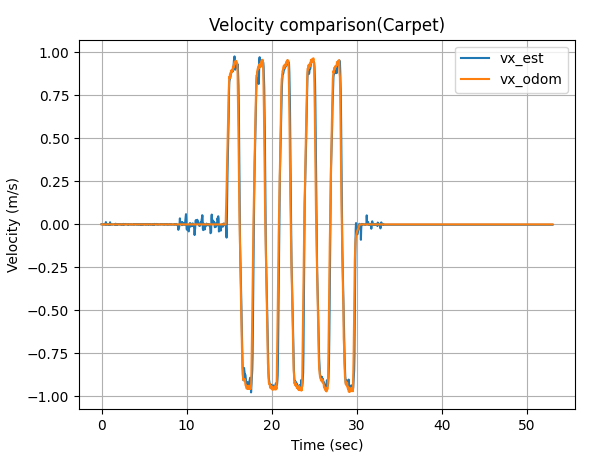}
\end{subfigure}
\hfill
\begin{subfigure}{0.48\columnwidth}
    \centering
    \includegraphics[width=\linewidth]{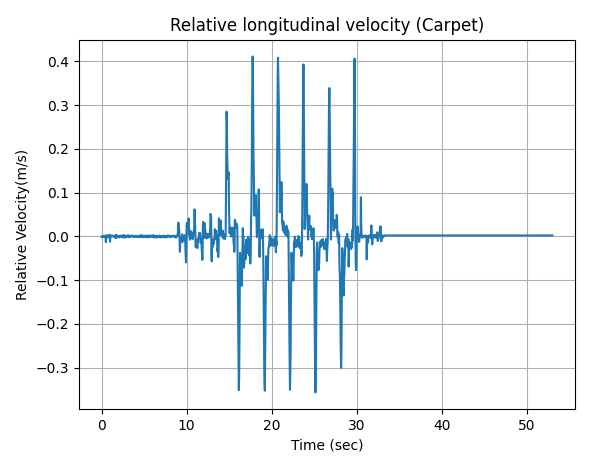}
\end{subfigure}
    \caption{Velocity comparison and relative velocity profile for Carpet surface.}
    \label{fig:Carpet_RV}
\end{figure}

Figures~\ref{fig:Concrete}--\ref{fig:Carpet_RV} show the velocity comparison between the factor graph estimated robot velocity and the wheel odometry-derived velocity across all five surfaces. They also show the relative velocity profiles for each of the five tested surfaces. Figure ~\ref{fig:Concrete} refers to one of the experiments conducted on the concrete surface in the MoCap facility.

A vinyl sheet has been firmly duct taped to the concrete surface and a series of experiments was conducted. Figure ~\ref{fig:Vinyl} shows the velocity comparison between the FG estimated velocity and the wheel odometry-derived velocity, and the relative longitudinal velocity of the robot.

A melamine board of thickness was duct taped to the surface, similar to the vinyl sheet and the data was collected. An appropriate time duration was chosen to ensure that the robot stays on the surface for the entire run, for all the surfaces. Figure ~\ref{fig:Board} shows the velocity comparison and relative velocity plots on this surface for one of the runs.

A 1:1 mixture of body wash and baby oil was applied to the concrete surface to create a slippery surface. Consequently, more slip was observed in this case compared to the rest of the surfaces. Figure ~\ref{fig:Soap} shows the same.

A high-performance polyester textured indoor carpet was used for a relatively high-traction surface. The robot slid the least on this surface of all five surfaces. Figure ~\ref{fig:Carpet_RV} shows the velocity profiles on this surface.
The velocity comparison plots in the Figures ~\ref{fig:Concrete}--\ref{fig:Carpet_RV} show a clear difference between low traction and high traction surfaces. For carpet, the estimated body velocity almost overlaps with the wheel velocity, while for the soap surface, the mismatch between the two velocities is largest among the five surfaces. This is also reflected in the relative velocity plots of the five surfaces, in both magnitude and slope of the up and down segments. For carpet, the rise and fall of relative velocity is much steeper compared to that of the soap surface. That is, the robot body velocity tracks the wheel velocity profile well. The velocity profiles show differences in both the magnitude and timing of the wheel-to-body mismatch. To examine how this mismatch varies within each transition, the relative-velocity signal was also analyzed in the frequency domain.
\subsection{Spectral analysis results}
\begin{figure*}[t]
    \centering
    \includegraphics[width=0.82\textwidth]
    {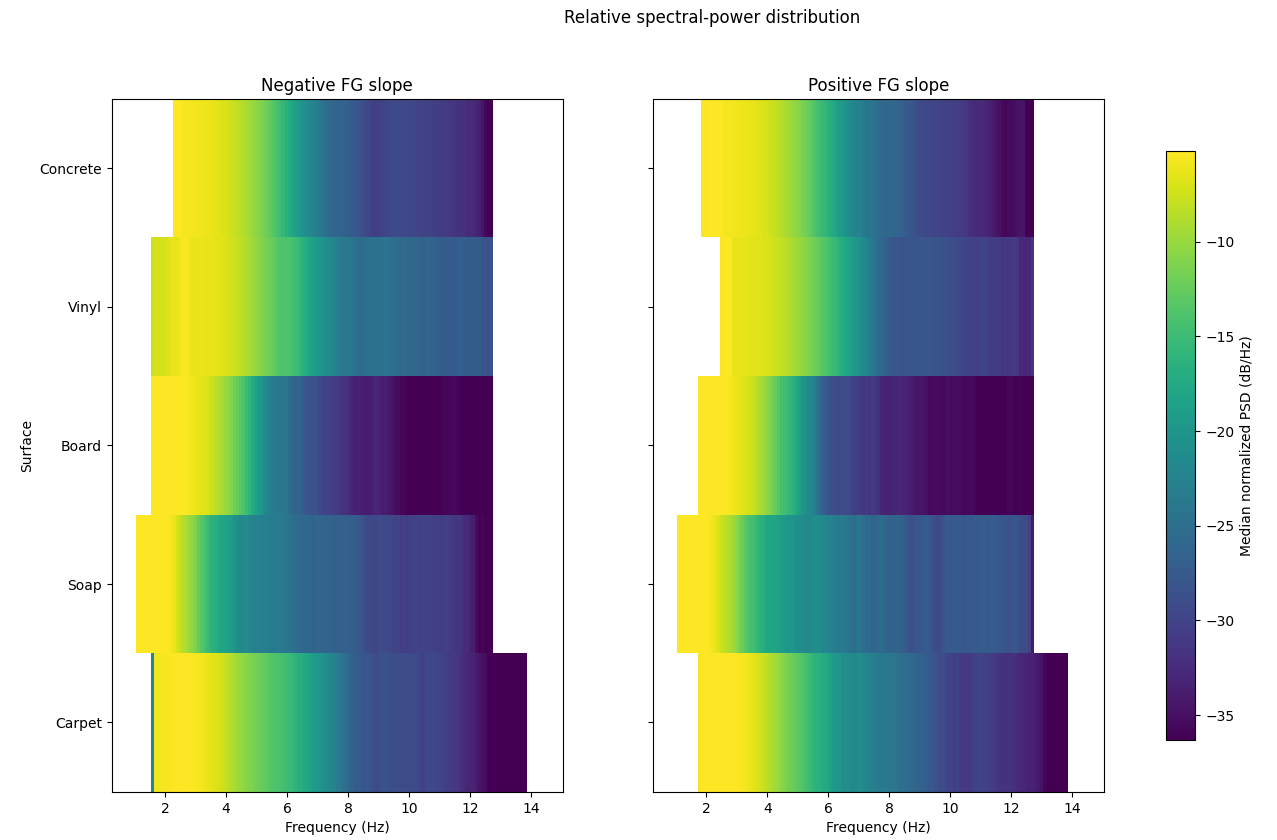}
    \caption{Normalized spectral-power distribution of the relative
    velocity during negative and positive FG slopes. Each segment was
    normalized by its total spectral power before the frequency-wise
    median was calculated. The heatmap therefore represents the
    distribution of mismatch power across frequency rather than its
    absolute magnitude. White regions correspond to frequencies
    excluded because of segment-resolution or sampling limitations.}
    \label{fig:normalized_relative_psd}
\end{figure*}
Figure~\ref{fig:normalized_relative_psd} shows that the
relative-velocity power was concentrated toward the lower portion of
the resolved frequency range for all surfaces. Concrete and Board
exhibited the most rapid reduction in normalized power with increasing
frequency, indicating that their velocity mismatch was primarily
concentrated at lower frequencies. Soap displayed a broader and flatter
spectral distribution, with a greater fraction of its power extending
toward the upper portion of the resolved band. Vinyl and Carpet showed
intermediate spectral distributions. Similar general patterns were
observed during positive and negative FG slopes.

The broader distribution observed for Soap indicates that its wheel-to-body velocity mismatch contained a greater contribution from faster fluctuations within transitions. However, the normalized spectrum contains no information about the total magnitude of the velocity mismatch and should not be interpreted independently as a ranking of surface friction. Instead, it demonstrates that the temporal structure of the mismatch differed between surfaces.
\begin{figure*}[t]
    \centering

    \subfloat[Two-dimensional LDA projection of the six run-level
    spectral features.]{
        \includegraphics[width=0.45\textwidth]
        {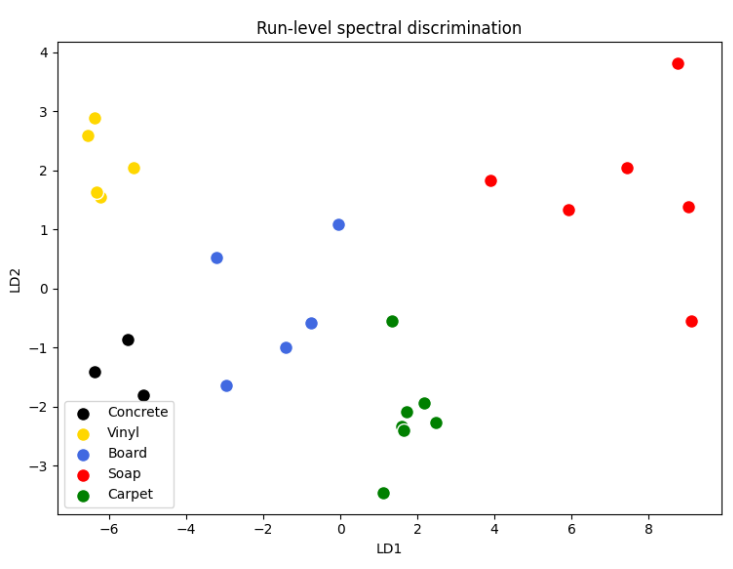}
        \label{fig:spectral_lda}
    }
    \hfill
    \subfloat[Confusion matrix obtained using leave-one-recording-out
    cross-validation.]{
        \includegraphics[width=0.45\textwidth]
        {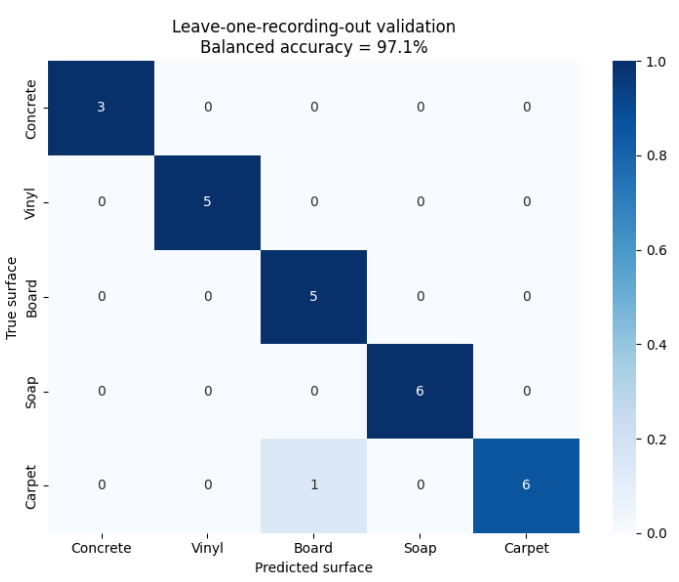}
        \label{fig:spectral_confusion}
    }

    \caption{Surface discrimination using run-level spectral features.
    Each point in (a) represents one complete recording. The LDA
    projection is a supervised visualization of the spectral-feature
    space. In (b), the matrix entries show the number of held-out
    recordings assigned to each surface, while the shading represents
    the row-normalized classification fraction.}
    \label{fig:spectral_discrimination}
\end{figure*}

\subsection{AUC analysis}
\begin{figure}[htb!]
    \centering
    \includegraphics[width=\columnwidth]{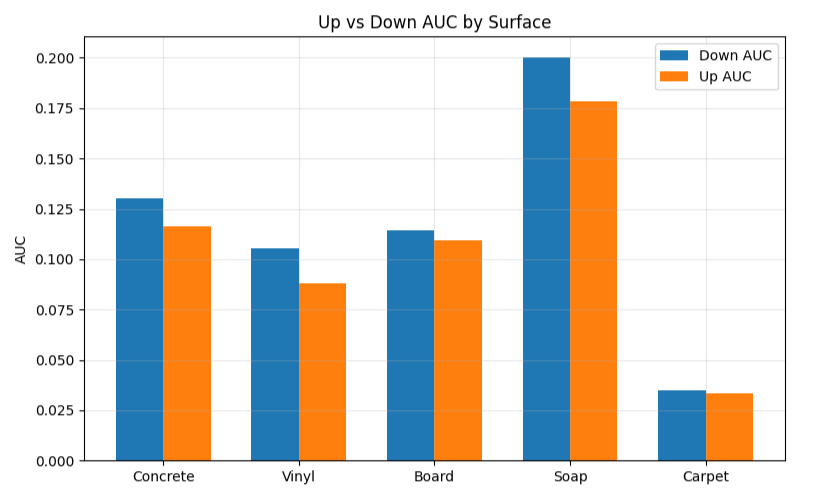}
    \caption{Median AUC for different surface relative velocity profiles for up and down segments.}
    \label{fig:AUC}
\end{figure}

\begin{table}[t]
\centering
\caption{Median static CoF values across surfaces measured by a drag-sled tribometer.}
\label{tab:tribometer}
\begin{tabular}{p{0.55\columnwidth} p{0.35\columnwidth}}
\toprule
\textbf{Surface} & \textbf{Median $\pm$ Std.} \\
\midrule
Board    & $0.420 \pm 0.080$ \\
Carpet   & $0.845 \pm 0.065$ \\
Concrete & $0.390 \pm 0.033$ \\
Soap     & $0.150 \pm 0.080$ \\
Vinyl    & $0.500 \pm 0.043$ \\
\bottomrule
\end{tabular}
\end{table}

The trend of varying surface slipperiness is clearly captured even in the AUC values of Fig.~\ref{fig:AUC}. The carpet has the lowest median AUC values for both up and down segments, the soap has the highest median AUC values, while the vinyl, board, and concrete fall in between, in that order. This trend suggests that robot wheels slipped the most in soap, followed by concrete, melamine board, and vinyl, and the least on carpet. The trend from the tribometer readings as shown in the Table~\ref{tab:tribometer}, is $Carpet > Vinyl > Board > Concrete > Soap$, ordered by decreasing CoF. This ordering aligns with the trend obtained from the wheel slip analysis. The distinction is clear between low-traction and high-traction surfaces, and thus, this method can be effective at detecting slippery surfaces, which can be used for correcting the wheel odometry drift caused by slip.

\begin{figure}
    \centering
    \includegraphics[width=\linewidth]{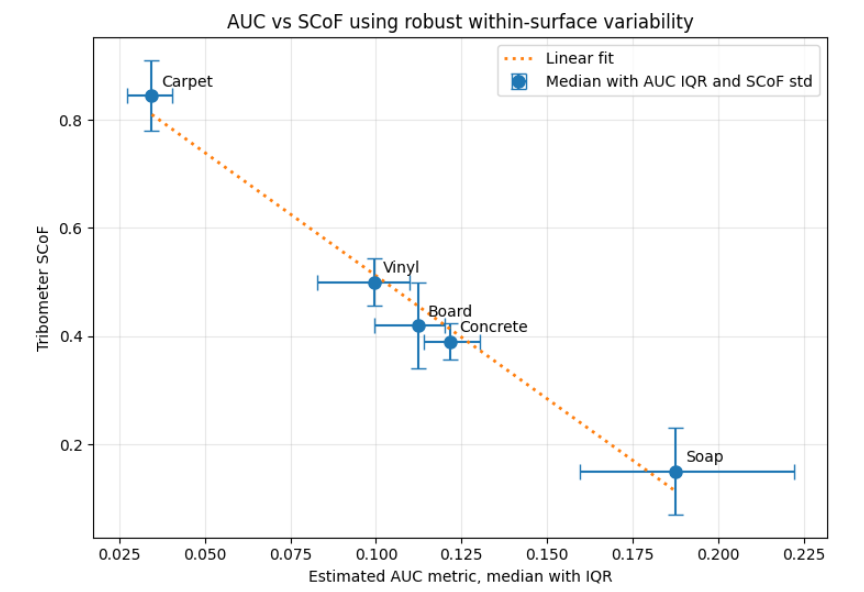}
    \caption{AUC vs SCoF; Horizontal error bars indicate the interquartile range (IQR) of the
estimated AUC values, while vertical error bars indicate the standard deviation of the tribometer
SCoF measurements.}
    \label{fig:AUCvsGT}
\end{figure}

The median AUC values for the up and down segments are averaged for each surface and plotted against their respective ground-truth friction values. Figure~\ref{fig:AUCvsGT} shows the inverse correlation between the median AUC values and the SCoF values measured by the tribometer. To quantify the relationship between the median AUC values and the tribometer measured static coefficient of friction (SCoF), a numpy polyfit first-order linear model was used, yielding Eq.~\eqref{eq_corr}. The resulting model was used to compute predicted coefficients of friction, $\mathrm{CoF}_{\mathrm{pred}}$, which were compared against the true values, $\mathrm{CoF}_{\mathrm{true}}$. The fit achieved $R^2 = 0.978$, indicating a strong negative linear correlation between AUC and SCoF for the tested surfaces. 

\begin{equation}
    \mathrm{CoF}_{\textrm{pred}} \approx 0.956 - 4.458\,*(\mathrm{AvgMedianAUC})
    \label{eq_corr}
\end{equation}
The AUC analysis captures the accumulated magnitude of the velocity mismatch, but does not distinguish a difference in amplitude from a delay in the body velocity response. The response-delay analysis examines this timing difference separately.

\subsection{Wheel-to-Body Velocity Response Delay}

\begin{figure*}[t]
    \centering
    \subfloat[Frequency-domain velocity response delay calculated from the phase 
    difference between the body and odometry velocities.]{
        \includegraphics[width=0.47\textwidth]
        {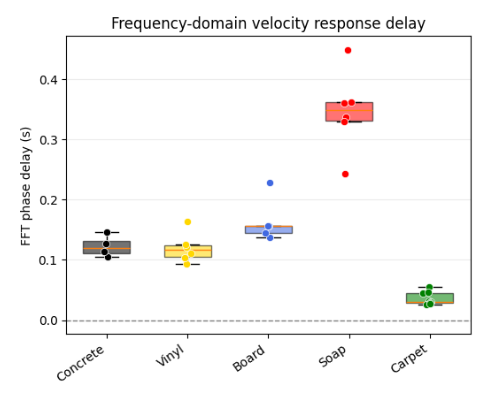}
        \label{fig:frequency_catchup}
    }
    \hfill
    \subfloat[Direct reversal velocity response delay calculated from the difference 
    between the body and odometry zero-crossing times.]{
        \includegraphics[width=0.47\textwidth]
        {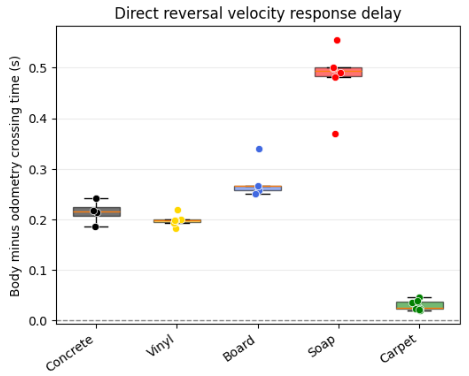}
        \label{fig:reversal_catchup}
    }
    \caption{Wheel-to-body velocity response delays measured for the five 
    surfaces. Each point represents one complete recording. The boxes show 
    the interquartile range, the center lines represent the medians, and the 
    whiskers extend to 1.5 times the interquartile range. A positive delay 
    indicates that the ground-referenced body velocity responded after the 
    wheel-odometry velocity.}
    \label{fig:catchup_delays}
\end{figure*}

\begin{figure}
    \centering
    \includegraphics[width=1\linewidth]{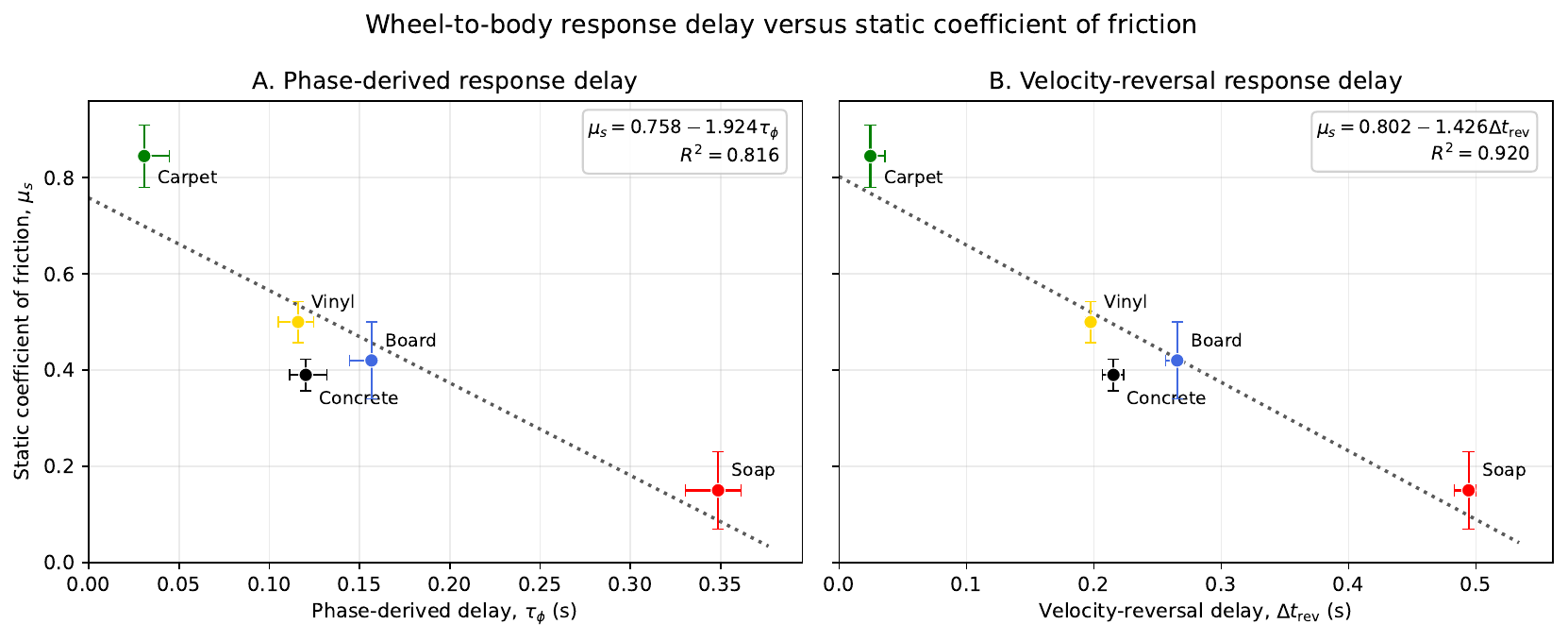}
    \caption{Relationship between the static coefficient of friction and (A) the phase-derived wheel-to-body response delay and (B) the velocity-reversal response delay. Each point represents the median delay across complete recordings for one surface. Horizontal error bars indicate the interquartile range of the response delay, while vertical error bars indicate the standard deviation of the measured static coefficient of friction. Dotted lines represent unweighted first-order linear fits.}
    \label{fig:Vel_res_vs_scof}
\end{figure}

Figure~\ref{fig:catchup_delays} compares the temporal response of the robot
body to changes in wheel-odometry velocity. The frequency-domain velocity response
delay, shown in Fig.~\ref{fig:frequency_catchup}, represents the effective
phase delay over the complete periodic maneuver. The direct reversal
catch-up delay, shown in Fig.~\ref{fig:reversal_catchup}, measures the time
between the odometry and body-velocity zero crossings during each change
in direction. All median delays were positive, indicating that the body
velocity generally responded after the wheel-odometry velocity.

Soap produced the largest delay using both methods. Its median
frequency-domain delay was approximately $0.349~\mathrm{s}$, compared with
$0.157~\mathrm{s}$ for Board, $0.120~\mathrm{s}$ for Concrete,
$0.116~\mathrm{s}$ for Vinyl, and $0.031~\mathrm{s}$ for Carpet. The direct
reversal analysis showed greater absolute delays but the same general
surface ordering. The median direct reversal delay was approximately
$0.494~\mathrm{s}$ for Soap, $0.265~\mathrm{s}$ for Board,
$0.215~\mathrm{s}$ for Concrete, $0.197~\mathrm{s}$ for Vinyl, and
$0.025~\mathrm{s}$ for Carpet. The difference in magnitude between the two
delay measures is expected because the frequency-domain metric represents
an effective delay over the full oscillatory response, whereas the direct
reversal metric focuses specifically on the time at which each signal
changes direction.

The large and comparatively variable delays measured on Soap indicate
that the wheel response was transferred to the robot body more slowly and
less consistently on this surface. This behavior is consistent with
temporary wheel-to-body decoupling during direction changes and supports
the identification of Soap as the surface exhibiting the strongest slip response. In contrast, the small delays observed on the carpet show that the body and wheel velocities changed direction almost
simultaneously. Concrete and vinyl surfaces exhibited similar intermediate delays, with substantial overlap between their recording-level measurements. Consequently, the present delay measurements do not provide strong evidence for separating concrete and vinyl surfaces based on temporal response alone.

Board surface exhibited the second-largest delay in both analyses, although its AUC-based velocity-mismatch response was not correspondingly large. This difference demonstrates that accumulated velocity mismatch and temporal delay describe separate aspects of wheel-to-body coupling. A recording
may contain a repeatable phase delay without producing a large or
sustained velocity difference. Therefore, the Board result should not be
interpreted as direct evidence that board was the second-most slippery
surface. Instead, it suggests that the response on board contained a
systematic timing component that may include vehicle inertia, controller response, signal filtering, or sensor synchronization.

Overall, the concurrence between the two independently calculated delay
metrics supports the conclusion that the measured differences were not
created solely by the frequency-domain calculation. Nevertheless, these
results represent a ranking of wheel-to-body response delay rather than a
direct ranking of surface friction. The command-reversal frequencies were
not identical across all surface experiments, and both phase delay and
zero-crossing delay can depend on the imposed acceleration and reversal
period. A direct friction comparison would therefore require identical
command waveforms and sensor-processing conditions for every surface.
Within the present dataset, the most defensible conclusion is that Soap
produced the clearest and largest reduction in the speed with which body
motion followed the wheel response.

\section{CONCLUSION AND FUTURE WORK}
Techniques were developed and compared to examine how the magnitude and timing of the wheel-to-body velocity mismatch relate to measured tire-floor friction. The results suggest that these measurements could help robots assess floor traction and identify slippery surfaces. This method can be easily extended by replacing motion capture updates with exteroceptive sensors like visual or LiDAR-based odometry. This approach can also be extended to incorporate lateral slip, enabling richer characterization of tire-surface interaction and improving the fidelity of floor slipperiness assessment. Furthermore, incorporating these slip estimation and surface characterization methods into state estimation could potentially be used to improve robot localization accuracy and robotic control. Similar approaches could also support slip-aware dead reckoning in outdoor environments with heterogeneous terrain conditions.





\section*{ACKNOWLEDGMENT}
We acknowledge the use of generative LLM (ChatGPT v.5.2) for editorial assistance as well as developing parts of the methodology.
\balance


\end{document}